# Maritime situational awareness using adaptive multi-sensor management under hazy conditions

Dilip K. Prasad[*], C. Krishna Prasath[*], Deepu Rajan[†], Lily Rachmawati[‡], Eshan Rajabally[§] and Chai Quek[†]

[*]Rolls-Royce@NTU Corporate Lab, Singapore

[†]School of Computer Science and Engineering, Nanyang Technological University, Singapore

[‡]Rolls-Royce Plc, Singapore

[§]Rolls-Royce Derby, United Kingdom

**Abstract:**

This paper presents a multi-sensor architecture with an adaptive multi-sensor management system suitable for control and navigation of autonomous maritime vessels in hazy and poor-visibility conditions. This architecture resides in the autonomous maritime vessels. It augments the data from on-board imaging sensors and weather sensors with the AIS data and weather data from sensors on other vessels and the on-shore vessel traffic surveillance system. The combined data is analyzed using computational intelligence and data analytics to determine suitable course of action while utilizing historically learnt knowledge and performing live learning from the current situation. Such framework is expected to be useful in diverse weather conditions and shall be a useful architecture to provide autonomy to maritime vessels.

## I. Introduction:

Industrial and academic research towards realization of autonomous maritime vessels (AMVs) is in full swing [1]-[28]. Several recent research works different aspects of this future technology. Examples include decision support [25]-[28], path planning [14]-[19], maritime image processing and computer vision [1]-[8], and control and regulations [9]-[13]. Hazy conditions pose significant risk for maritime traffic management at busy ports. Recent history of persistent haze in South East Asia clearly indicates the need of maritime traffic systems being adaptable and robust to sudden onset of haze and differing haze levels [29], [30]. Nevertheless, poor visibility conditions have always posed challenges, even in the current advanced maritime systems which still rely on human experience while providing all sensor support possible by the current technology. In the autonomous maritime technology, with almost close to zero-human-intervention, difficult weather conditions pose a more severe challenge.

Since the sensor technology is quite mature and no major innovation in sensors is expected, the onus of poor-weather autonomy lies upon designing systems that integrates the existing sensors, performs intelligent analyses of their combined data to generate as rich situational awareness as possible, and determines appropriate action plan for autonomy [22]-[27], [31]. It is well known that radar, electro-optical visible range cameras, and electro-optical infrared cameras provide complimentary information due to different propagation characteristics [32], nature and kind of information, and specific merits and demerits. Further, weather sensors (rain sensors, aerosol particle sensors, humidity sensors, luminance sensors, etc.) and geo-sensors (global positioning system, compass, gyroscopes, etc.) augment the situation awareness [25], [27]. This paper presents a multi-sensor architecture with an adaptive multi-sensor management system that can provide such a support for poor-visibility conditions. The architecture is discussed in detail in section 2. The paper is concluded in section 3.

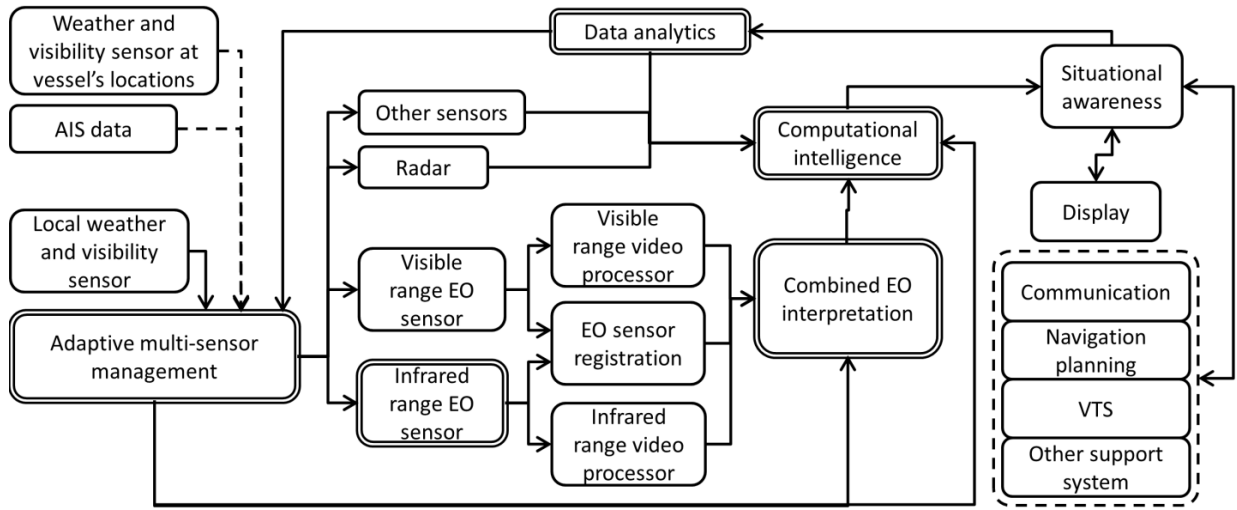

**Fig 1: Proposed adaptive multi-sensor management architecture. Blocks with double line boundaries are critical to the architecture.**

## II.     Multi-sensor architecture

The block diagram of the architecture is shown in Fig. 1. The system controls the sensors and uses computational intelligence based on data analytics to combine the sensors' data interpretations appropriately and adaptively, subject to the weather conditions. The blocks with double line boundaries in Fig. 1 are critical to maritime situational awareness in hazy conditions. The adaptive multi-sensor management block controls the sensors and provides relevant information to the computational blocks. It takes input from various local and remote sensors and decision and planning modules. Infrared block is selectively used and positioned to complement the visible range EO sensor subject to hazy conditions and scene coverage requirements. Combined EO interpretation not only presents combined graphical information from both EO sensors, it also determines how much, how reliably, and what kind of EO interpretation should be derived from both the sensors given the weather conditions and port requirements. Computational intelligence block performs intelligent combination of active data generated from the sensors to lead to the current situational awareness and the data analytics block builds the long terms learning and intelligence of the system for robust operation.

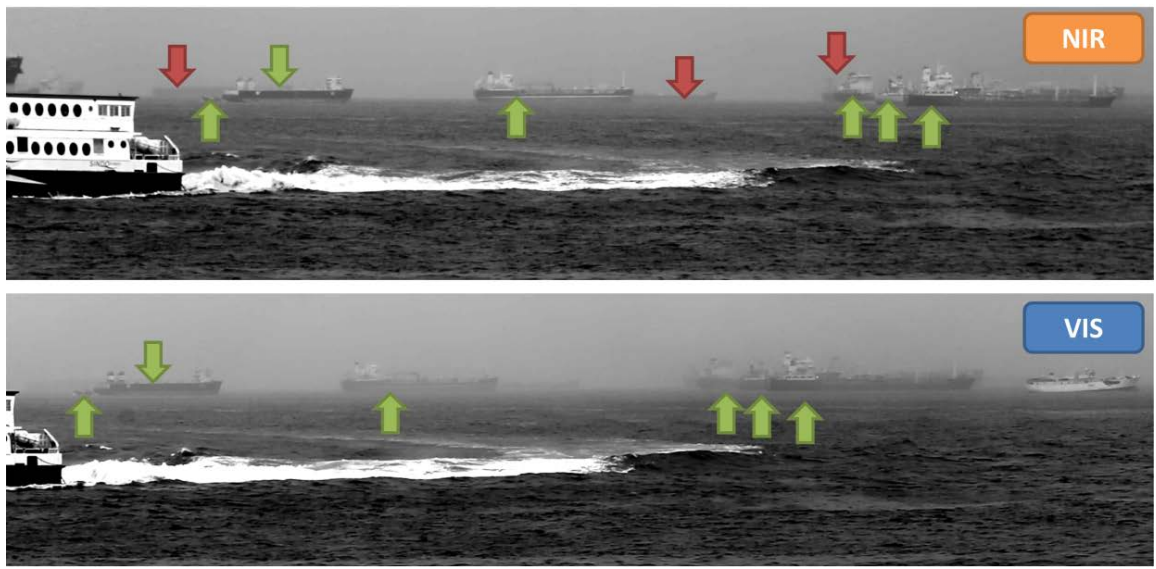

**Fig 2: An example of the advantage of near infrared imaging over visible range imaging in hazy conditions.**

## A.    Information from sensors

In this section, we discuss the on-board and remote sensors used in the proposed architecture. These sensors include sensors mounted on-board the AMV, the sensors mounted on other vessels such that they function as remote sensors for the AMV, and the information from the on-shore vessel traffic surveillance system (if any) in communication with the AMV. They are discussed in the following sub-sections.

### 1.    *On-board sensors*

The on-board sensors can be divided into three broad classes, namely imaging sensors, weather and geo-sensors, and vessels' health sensors. The imaging sensors include radar sensors, electro-optical (EO) visible cameras, EO infrared (IR) cameras, and sonars. Electro-optical sensors may be monocular, stereoscopic [33], or have other specific functions. EO visible cameras may further be color cameras or intensity cameras. EO-IR cameras may be long-wave IR (such as FLIR [34]), mid-wave IR [35], or near IR [36]. There are typically multiple EO sensors to allow sufficient coverage of scene around the AMV. There is typically at least one radar on-board any medium or large size maritime vessel. Most commercial ships are equipped with sonars as well. Each of these sensors provide an image of the part of AMV's scene that it covers. They each have their benefits as well limitations. For example, radar provides locations of vessels or land features within distances of few kilometers to few hundred kilometers from the AMVs. Radars are quite accurate in locating vessels with large radar signatures and provide all weather imaging. However, small vessels may be missed or miss-located by radars. Further, their resolution is few meters and they have a shadow reason close to AMVs such that vessels or features closer than a few kilometers cannot be detected or tracked by radars. Similar consideration apply to sonars as well which provide imaging in water and help in detecting underwater obstacles.

The shadow region of radar close to AMV and the criticality of scene imaging close to AMVs for avoiding collision at close quarters necessitate the use of EO sensors. As noted before, visible range EO sensors are severely afflicted by poor weather conditions, particularly by haze, and are not suitable for night-vision. Thus, infrared EO sensors are important for poor-visibility conditions and night vision. For example, consider the visible (VIS) and near infrared (NIR) images of the same maritime scene taken on a hazy day in Singapore. The red arrows in Fig. 2 show the vessels that are completely invisible in VIS image but decipherable in NIR image. Further, while generally the sharpness and color information in visible range is of merit in usual daylight conditions, the sharpness (see Fig. 2) and color information is significantly compromised on a hazy day and subject to the haze measurements, infrared data becomes increasingly more valuable.

Compasses, gyroscopes, global positioning systems, velocity sensors, etc. serve as geo-sensors that provide information of geological position and track of the vessel. Further, wind sensors, humidity sensors, aerosol particulate sensors, luminance sensors, tidal sensors, water temperature etc. comprise the weather sensors. Geo-sensors and weather sensors help in planning navigation course of the vessel. Among these, aerosol particulate measurement and luminance measurement are direct indicators of the visibility conditions. In the current technology, they are typically used by trained personnel to make navigation plans while computational means are used to simply display the data or provide course projections based on the current scenario and chosen plan. However, in AMVs, their data has to be used for computational interpretation as well as decision making.

The third category of sensors are the vessel health sensors, which generate the diagnostic information about vessel's functional and stand-by modules, such as rudder speed, engine temperature, and so on. While these sensors do not play a critical role in situational awareness (external), deviation from their normative values are important considerations in navigation planning.

## 2. *Sensor information from other vessels and VTS*

Automatic identification system (AIS) is now compulsory on all commercial vessels. It is the information that is broadcast by a vessel to inform other vessel and onshore facilities about its presence and current path. It includes identifier for the vessel, its geo-location and current navigation path, speed, etc. The information may be received by other vessels through very high frequency (VHF) radio waves or through satellite communication. For use in AMVs, the AIS data can be augmented with the data of weather sensors at the locations of the other vessels, such that a crude spatial map of the weather conditions may be obtained for a better situation awareness in terms of weather. Further, if the AMV is communication with an on-shore vessel traffic surveillance system, then it can receive more information about navigation traffic and conditions.

B.  Adaptive multi-sensor management

The adaptive multi-sensor management block is a primary intelligence block of the architecture which analyses all the non-imaging sensor data and generates an assessment of the weather conditions. Subsequently, it uses the assessment to manage the imaging sensors adaptively as well as to determine the frequency of monitoring of each sensor. It takes inputs from all local and remote sensors except the imaging sensor.

It also takes input from data analytics block which provides previously learnt references for weather conditions. Further, it generates functional information for all the local sensors, EO interpretation block, and computational intelligence block. The internal functioning of this block is heavily dependent upon the data analytics block. It uses the weather dictionary of the data-analytics block, which would contain detection and follow-up data for each discriminable weather condition. It is desirable to make the adaptive multi-sensor management block auto-programmable to a certain extent, or at least have pre-programmed libraries of programs that can be loaded depending upon the current recommendations of always evolving data analytics block. Further, it should have a certain memory allocated for each sensor to store recent history of the sensor data. Further, it needs memory to store recent history of weather. The detection of weather conditions would require incremental learning and detection approach []. If the recent history of weather does not indicate change in weather conditions, the weather detection template for the previously logged weather condition is used, statistical features in this template are computed using the sensors' historical and current data and check for deviation from the previously logged weather condition is performed. If a deviation is detected, depending upon the nature of deviation, more templates from the weather dictionary in the data analytics block are loaded and the actual weather condition determined. If there is no proper match to the weather templates in the weather dictionary, i.e. the weather condition is new to the data analytics block, a linear combination of the closest weather conditions is used and computational intelligence block is intimated about the newly registered weather condition.

Having detected the weather conditions, it initiates the following actions. First, it determines the frequency of logging measurements from the non-imaging sensors. For example, if the weather conditions have consistently been sunny and dry, the frequency of querying the humidity and aerosol particulate sensors may be reduced. However, if rain is considered imminent or hazy conditions are present, the corresponding sensors need to be queried more often but luminance sensor need not be queried frequently. Second, it determines the controls of the imaging sensors and the weightage of combining the imaging data from the imaging sensors. For example, in sunny and clear conditions, the EO visible sensor and the radar are the most reliable sensors. The radar calibration corresponding to the sunny weather conditions may be used. Further, the electronic settings of the EO visible sensors can be appropriately adjusted, for example large dynamic range, focus at long distance, electronic gain of about one, vivid color settings, etc. On the other hand, in hazy conditions, radars, EO visible, and EO infrared sensors are all needed. At a close range, EO visible sensor with small dynamic range, large gamma correction, close focus is sufficient and EO visible sensor can be given maximum weightage for the first few meters. For distances larger than few meters to few kilometers, the maximum weight should be given to the EO-IR sensors with long focus and appropriate gamma correction. For regions beyond, radar has to be given the maximum weightage. Further, in either weather conditions, AIS data from other vessels and VTS to be incorporated in situation awareness has to be passed to the computational intelligence block with highest weightage, as they can be deemed as extremely reliable. The adaptive multi-sensor management block directly determines and controls the settings of the sensor. It passes the AIS data and the weightage of combination of different sensors to the computational analytics block. It also passes the weightage information to the combined EO interpretation block which pre-combines the EO visible and EO-IR data and uses computer vision techniques to segment and map the foreground vessels.

## C.     EO sensor processing and combined EO interpretation

EO sensor processing is done across several blocks including one EO visible range video processor per EO visible sensor, one EO infrared range video processor per EO-IR sensor, a single EO sensor registration block for all EO sensors and a single combined EO interpretation block for all EO sensors. All the blocks for EO sensor processing use techniques of image processing, computer vision, and machine learning.

The visible and infrared video processors, one for each sensor, perform sensor specific operations such as distortion correction, intensity scaling, contrast enhancement, fiducial marking, filtering and noise suppression etc. Further, some pre-learning such as extraction of color features, statistical features, geometric features, patch features, etc. can be done for each sensor independently in a more efficient manner.

The corrected video feeds, their fiducial markings, local features are then passed to the EO sensor registration block which uses fiducial marking and the stored information about the sensor coordinates relative to sensor coordinates to register each video feed and corresponding features in the world coordinates. Further, all the videos and features are transformed from their internal pixel and frames units to physical world units, such as meters, seconds, meters/second, radians, etc.

The combined EO interpretation block uses computer vision algorithms [4] with suitable weighted features from all the video feeds (weights determined by the adaptive multi-sensor management block) to suppress the dynamic background of water, extract the foreground vessels, small objects, and terrestrial structures of interest. It also tracks the mobile objects. Further, it classifies all detected objects according to their sizes, models, distances from the AMV, motion patterns, etc. The combined EO interpretation block may store the feature knowledge and the trained classifiers either locally or may store and evolve them in the data analytics block. In our opinion, it is better to keep data analytics block corresponding to situational, weather, and sensor management separate and if needed a dedicated computer vision data analytics block may be included in the EO processing blocks. We also note that this block does uses the pre-computed features of the video feeds more than the video feeds themselves in order to reduce the computational load entailing the video processing. However, video feed is retained for display and projection purposes.

### D. Computational intelligence

Computational intelligence block functions as the cognitive brain of the architecture. It receives processed results from all the functional blocks, namely the adaptive multi-sensor management block, the combined EO interpretation block, the radar and sonar post-processed results, AIS data and other relevant sensors. It generates at least three types of awareness, namely navigational situational awareness, weather situational awareness, and need-to-learn awareness.

It first super-imposes the AIS data, radar and sonar results, and the EO output and identifies one-to-one correspondences between these. In case of some loose correspondences, it uses fuzzy logic to identify suitable correspondences. Further, for the items without any correspondences, it uses fuzzy logic to classify them as spurious (likely false alarms), small objects (which may be unequipped with AIS, may be undetectable by radar or may be in the shadow region of radar), or fixed terrestrial or oceanic structures such as islands or icebergs, or rigs, which may be friendly or threatening. Then, the motion patterns and expected paths of the various objects, including the path of the AMV, are super-imposed. Then, high risk objects and situations are identified. High risk object may be small sized vehicles (potentially without AIS), fast moving vessels, vessels performing complex maneuvers, vessels very close to AMV, icebergs, vessels with paths intersecting with AMV's path. Temporal projections are performed to determine the sequence of events. These generate navigational situational awareness.

The weather sensor data of the AMV and the other vessels are used to understand the current and persisting weather conditions and predict the weather conditions in the vicinity and in the near-future. For example, it may analyze the data of local and remote humidity sensors, rain sensors, wind sensors, etc. to predict the strength, direction, disposition, and duration of a storm. It may also find pocket regions which are relatively less affected by storm, thus aiding navigation. Or it may identify the direction of winds laden by particulate causing haze, which may help the AMV navigate to a better visibility zone.

Notably, in both the above awareness, the computational intelligence block draws heavily from the data analytics block which learns patterns and suitable course of actions from weather or navigation events, event markers, and results of actions undertaken in various events. Even while using this knowledge from the data analytics block, the computational intelligence block also generates the need-to-learn awareness. In this awareness, the computational block identifies the frequency or occurrence of certain events over time and decides if the data analytics block needs to reinforce learning of such events. If a rare or new event weather or navigational event occurs, the computational intelligence block uses current data analytics to proceed with fuzzy situational awareness but generate a high need-to-learn awareness, which then informs the data analytics block to register and learn the features of the event from the raw data of sensors, intermediate features, and fuzzy situational awareness.

### E.    Situation awareness and course of action

The computational analysis block, having performed situation awareness (both navigational and weather) needs to inform about the situation to the decision-making blocks, communication blocks, and information tracking blocks. These blocks may be navigation plan and update block, an on-shore vessel traffic surveillance system, ATS broadcast by the AMV, express communication to other specific vessels for collision avoidance, and display for any on-board or remote human crew.

### F.    Data Analytics

The function of data analytics block has been discussed in a dispersed manner in the above sub-sections already. Data analytics block performs learning and memorizing of the knowledge developed over time by the AMV and provides the knowledge in usable form to different blocks.

Data analytics block learns specifically about situations, conditions, and specific events concerning weather and navigation. Situation refers to state of weather (rainy, sunny and clear, hazy, windy, stormy, etc.) or state of navigation (keeping through, performing certain maneuver, coordinating movements with certain vessel). Condition refers to the features or characteristics indicating the situation. For example, high wind speeds, heavy rains, poor luminance, etc. characterize stormy conditions. Events are specific situations which necessitate a review of course of action. They may include detection of collision threat, imminent storm, sudden deterioration of weather conditions, etc. Learning is performed in tandem with the need-to-learn awareness generated by the computational intelligence block. It either reinforces the learnt situations and their conditions, or updates the learnt situations with new conditions additionally characterizing it, or registers new events and their features. It may also register or learn the features of the course of action taken in different situations. This is of critical importance upon occurrence of events and may serve as a general guide in other routine situations. It would make use of diverse machine learning techniques such as boosting, classifier trees, bag of words, feature dictionaries, fuzzy neural networks, etc.

It provides weather templates from the weather dictionary in a ranked manner to the adaptive multi-sensor management block. The most recently used weather templates rank the highest and are first provided, most recently reinforced or updated weather templates follow them, events related to either the most recently used or the most recently updated templates appear after these. All other weather situations have the lowest rank. However, the rank in not in the form of list, but in the form of a connected state network, where each node is a weather situation and its rank is its nodal weight. The directed nodes connected to it represent possible subsequent state. Thus, the problem of supplying weather templates to the adaptive multi-sensor management block is cast as a Markov random process although the learning of each weather situation is not through Markov processes. Each template contains the weather situation, the weather sensors' features that characterize it and are used for detecting the weather condition, and the suggested weightage and frequency of sensors if the situation is detected.

It also supplies weather and navigation situational knowledge to the computational intelligence block. For example, it provides weather and its forecast options to the computational intelligence block where the above-mentioned weather Markov random process is used to identify likely forecasts. Or, if an event is detected (imminent or currently on-going), say a storm, it suggests looking for safe pocket regions. Or, if there is a geo-location conflict between different sensors regarding the same vessel, it provides pre-learnt conflict resolution templates (fuzzy logic) depending upon the nature and magnitude of conflict. Or, given the motion pattern, location, and type of a vessel, it provides template for potential threat assessment. In short, it serves the function of knowledge curator and consultant to the blocks of this architecture

## III. Conclusion

We have presented a multi-sensor architecture for weather-robust situational awareness suitable for autonomous maritime vessels of future, which is an adaptive, automatic, self-evolving system of knowledge based sensor management and intelligence generation. We discuss the requirements, functions, and capabilities of each block and highlight suitable technologies or techniques, where available. We acknowledge that this architecture is a futuristic and multi-technology architecture and although its technical feasibility is undoubtable, its implementation and fine-tuning may require a few more years of research and development by a team of research scientists. Regardless, this architecture is a major step towards the fruition of AMV technology.

## IV. References:

[1] Kristan, Matej, et al. "Fast image-based obstacle detection from unmanned surface vehicles." IEEE transactions on cybernetics 46.3 (2016): 641-654.
[2] D. K. Prasad, C.K. Prasath, D. Rajan, C. Quek, L. Rachmawati, and  E. Rajabally, "Challenges in video based object detection in maritime scenario using computer vision," 19th International Conference on Connected Vehicles, Zurich, 13-14 January, 2017.

[3] Woo, Joohyun, and Nakwan Kim. "Vision based obstacle detection and collision risk estimation of an unmanned surface vehicle." Ubiquitous Robots and Ambient Intelligence (URAI), 2016 13th International Conference on. IEEE, 2016.

[4] D. K. Prasad, D. Rajan, L. Rachmawati, E. Rajabally, and C. Quek, "Video processing from electro-optical sensors for object detection and tracking in maritime environment: A Survey." IEEE Transactions on Intelligent Transportation Systems, vol. 18, (2017).

[5] Mentzelos, Konstantinos. Object localization and identification for autonomous operation of surface marine vehicles. Diss. Massachusetts Institute of Technology, 2016.

[6] D. K. Prasad, D. Rajan, L. Rachmawati, E. Rajabally, and C. Quek, "MuSCoWERT: multi-scale consistence of weighted edge Radon transform for horizon detection in maritime images," Journal of Optical Society America A, 33.12 (2016): 2491-2500.

[7] D. K. Prasad, D. Rajan, C. Krishna Prasath, L. Rachmawati, E. Rajabally, and C. Quek, "MSCM-LiFe: Multi-Scale Cross Modal Linear Feature for Horizon Detection in Maritime Images," IEEE TENCON, Singapore, 22-25 Nov, 2016.

[8] Sadhu, Tanmana. Obstacle detection for image-guided surface water navigation. Diss. 2016.

[9] Naeem, Wasif, Sable C. Henrique, and Liang Hu. "A Reactive COLREGs-Compliant Navigation Strategy for Autonomous Maritime Navigation." IFAC-PapersOnLine 49.23 (2016): 207-213.

[10] Porathe, Thomas, Hans-Christoph Burmeister, and Ørnulf Jan Rødseth. "Maritime unmanned navigation through intelligence in networks: The MUNIN Project." 12th International Conference on Computer and IT Applications in the Maritime Industries, COMPIT'13, Cortona, 15-17 April 2013. 2013.

[11] Sarda, Edoardo I., et al. "Station-keeping control of an unmanned surface vehicle exposed to current and wind disturbances." Ocean Engineering 127 (2016): 305-324.

[12] Qin, Zihe, Hanbing Sun, and Zhuang Lin. "A Hierarchical Control Strategy of Autonomous Unmanned Surface Vehicle Swarm." The 26th International Ocean and Polar Engineering Conference. International Society of Offshore and Polar Engineers, 2016.

[13] Woo, Joohyun, and Nakwan Kim. "Design of Guidance Law for Docking of Unmanned Surface Vehicle." Journal of Ocean Engineering and Technology 30.3 (2016): 208-213.

[14] Yunsheng, Fan, Sun Xiaojie, and Wang Guofeng. "On model parameters identification and fuzzy self-adaptive course control for USV." Control and Decision Conference (CCDC), 2016 Chinese. IEEE, 2016.

[15] Kim, Taejin, et al. "Development of a multi-purpose unmanned surface vehicle and simulation comparison of path tracking methods." Ubiquitous Robots and Ambient Intelligence (URAI), 2016 13th International Conference on. IEEE, 2016.

[16] Liu, Y. Path planning algorithms for unmanned surface vehicle formation in maritime environment. Diss. UCL (University College London), 2016.

[17] Zhu, Man, et al. "Maritime Unmanned Vehicle Cruise Path Planning for Maritime Information Collection." IFAC-PapersOnLine 49.23 (2016): 103-108.

[18] Niu, Hanlin, et al. "Efficient Path Planning Algorithms for Unmanned Surface Vehicle." IFAC-PapersOnLine 49.23 (2016): 121-126.

[19] Liu, Deli, et al. "Dynamic replanning algorithm of local trajectory for unmanned surface vehicle." Control Conference (CCC), 2016 35th Chinese. TCCT, 2016.

[20] Huang, Zhen, et al. "Design of an Intelligent Trimaran USV for Maritime Rescue." The 26th International Ocean and Polar Engineering Conference. International Society of Offshore and Polar Engineers, 2016.

[21] Zhang, Ru-Bo, Li-Qun Duan, and Chang-Ting Shi. "A method to evaluate autonomous performance for Unmanned Surface Vehicle." Control Conference (CCC), 2016 35th Chinese. TCCT, 2016.

[22] N. Pires, J. Guinet, and E. Dusch, "ASV: an innovative automatic system for maritime surveillance," Navigation 58, 1–20 (2010).

[23] Ince, A. Nejat, et al. Principles of integrated maritime surveillance systems. Vol. 527. Springer Science & Business Media, 2012.

[24] Ahvenjärvi, S. "The Human Element and Autonomous Ships." TransNav: International Journal on Marine Navigation and Safety of Sea Transportation 10 (2016).

[25] Han, Jungwook, et al. "GPS-less Coastal Navigation using Marine Radar for USV Operation." IFAC-PapersOnLine 49.23 (2016): 598-603.

[26] Pietrzykowski, Zbigniew, and Piotr Wołejsza. "Decision Support System in Marine Navigation." Challenge of Transport Telematics: 16th International Conference on Transport Systems Telematics, TST 2016, Katowice-Ustroń, Poland, March 16–19, 2016, Selected Papers. Springer International Publishing, 2016.

[27] Arias Medina, Daniel, et al. "Robust Position and Velocity Estimation Methods in Integrated Navigation Systems for Inland Water Applications." (2016).

[28] Lazarowska, Agnieszka. "A new deterministic approach in a decision support system for ship's trajectory planning." Expert Systems with Applications (2016).

[29] Quah, Euston, and Helena Varkkey. "The political economy of transboundary pollution: mitigation forest fires and haze in Southeast asia." The Asian Community: Its Concepts and Prospects (2013): 323-358.

[30] Salinas, Santo V., Astrid Muller, and Tan Li. "A comparative study of aerosol optical depth over Singapore From 2007-2014: Results from Aeronet and Modis data analysis."

[31] S. P. van den Broek, P. B. Schwering, K. D. Liem, and R. Schleijpen, "Persistent maritime surveillance using multi-sensor feature association and classification," in "SPIE Defense, Security, and Sensing," (2012),pp. 83920O–83920O.

[32] C. C. Chen, "Attenuation of electromagnetic radiation by haze, fog, clouds, and rain," Tech. rep., DTIC Document (1975).

[33] Huntsberger, Terry, et al. "Stereo vision–based navigation for autonomous surface vessels." Journal of Field Robotics 28.1 (2011): 3-18.

[34] MacIntyre, Iain. "FLIR Systems: Best thermal imaging supplier by a long shot." Ausmarine 35.3 (2013): 44.

[35] Krapels, Keith, et al. "Midwave infrared and visible sensor performance modeling: small craft identification discrimination criteria for maritime security." applied optics 46.30 (2007): 7345-7353.

[36] Smith, A. A., and M. K. Teal. "Identification and tracking of maritime objects in near-infrared image sequences for collision avoidance." Image Processing And Its Applications, 1999. Seventh International Conference on (Conf. Publ. No. 465). Vol. 1. IET, 1999.